%% file: ln_paper_arxiv.tex
\documentclass{article} 
\usepackage[final]{nips_2016}

\usepackage{hyperref}
\usepackage{url}
\usepackage{textcomp}
\usepackage{floatrow}
\usepackage{graphicx}
\usepackage{amsmath}
\usepackage{amsfonts}
\usepackage{todonotes}
\usepackage{subfigure}
\usepackage{epstopdf}
\usepackage{tabulary}
\usepackage{multirow}
\usepackage{color}
\usepackage{booktabs}
\usepackage{wrapfig}
\usepackage{bm}
\newfloatcommand{capbtabbox}{table}[][\FBwidth]

\graphicspath{ {figures/}}

\title{Layer Normalization}

\author{
Jimmy Lei Ba \\
University of Toronto \\
{\small \texttt{jimmy@psi.toronto.edu}} \\
\And
Jamie Ryan Kiros \\
University of Toronto \\
{\small \texttt{rkiros@cs.toronto.edu}} \\
\And
Geoffrey E. Hinton \\
University of Toronto \\
and Google Inc.\\
{\small \texttt{hinton@cs.toronto.edu} }
}

\input{macros.tex}

\begin{document}

\maketitle

\begin{abstract}

Training state-of-the-art, deep neural networks is computationally expensive. One way to reduce the training time is to normalize the activities of the neurons.  A recently introduced technique called batch normalization uses the distribution of the summed input to a neuron over a mini-batch of training cases to compute a mean and variance which are then used to normalize the summed input to that neuron on each training case.  This significantly reduces the training time in feed-forward neural networks. However, the effect of batch normalization is dependent on the mini-batch size and it is not obvious how to apply it to recurrent neural networks.  In this paper, we transpose batch normalization into layer normalization by computing the mean and variance used for normalization from all of the summed inputs to the neurons in a layer on a {\it single} training case.  Like batch normalization, we also give each neuron its own adaptive bias and gain which are applied after the normalization but before the non-linearity.  Unlike batch normalization,  layer normalization performs exactly the same computation at training and test times.  It is also straightforward to apply to recurrent neural networks by computing the normalization statistics separately at each time step.  Layer normalization is very effective at stabilizing the hidden state dynamics in recurrent networks. Empirically, we show that layer normalization can substantially reduce the training time compared with previously published techniques.    

\end{abstract}

\section{Introduction}
\label{sec:intro}

Deep neural networks trained with some version of Stochastic Gradient Descent have been shown to substantially outperform previous approaches on various supervised learning tasks in computer vision \citep{krizhevsky2012imagenet} and speech processing \citep{hinton2012deep}. But state-of-the-art deep neural networks often require many days of training. It is possible to speed-up the learning by computing gradients for different subsets of the training cases on different machines or splitting the neural network itself over many machines \citep{dean2012large}, but this can require a lot of communication and complex software.  It also tends to lead to rapidly diminishing returns as the degree of parallelization increases. An orthogonal approach is to modify the computations performed in the forward pass of the neural net to make learning easier. Recently, batch normalization \citep{ioffe2015batch} has been proposed to reduce training time by including additional normalization stages in deep neural networks. The normalization standardizes each summed input using its mean and its standard deviation across the training data. Feedforward neural networks trained using batch normalization converge faster even with simple SGD. In addition to training time improvement, the stochasticity from the batch statistics serves as a regularizer during training. 

Despite its simplicity, batch normalization requires running averages of the summed input statistics. In feed-forward networks with fixed depth, it is straightforward to store the statistics separately for each hidden layer. However, the summed inputs to the recurrent neurons in a recurrent neural network (RNN) often vary with the length of the sequence so applying batch normalization to RNNs appears to require different statistics for different time-steps. Furthermore, batch normalization cannot be applied to online learning tasks or to extremely large distributed models where the minibatches have to be small. 

This paper introduces layer normalization, a simple normalization method to improve the training speed for various neural network models. Unlike batch normalization, the proposed method directly estimates the normalization statistics from the summed inputs to the neurons within a hidden layer so the normalization does not introduce any new dependencies between training cases. We show that layer normalization works well for RNNs and improves both the training time and the generalization performance of several existing RNN models. 

\section{Background}
\label{sec:model}

A feed-forward neural network is a non-linear mapping from a input pattern $\data$ to an output vector $y$. Consider the $l^{th}$ hidden layer in a deep feed-forward, neural network, and let $\act^l$ be the vector representation of the summed inputs to the neurons in that layer. The summed inputs are computed through a linear projection with the weight matrix $\Weights^l$ and the bottom-up inputs $\hidden^l$ given as follows:
\begin{equation}
\act^l_i = {\weights^l_i}^\T \hidden^l
\qquad
\hidden^{l+1}_i = \actFunc(\act^l_i + \bias^l_i)
\end{equation}
where $\actFunc(\cdot)$ is an element-wise non-linear function and $\weights^l_i$ is the incoming weights to the $i^{th}$ hidden units and $\bias^l_i$ is the scalar bias parameter. The parameters in the neural network are learnt using gradient-based optimization algorithms with the gradients being computed by back-propagation. 

One of the challenges of deep learning is that the gradients with respect to the weights in one layer are highly dependent on the outputs of the neurons in the previous layer especially if these outputs change in a highly correlated way. 
Batch normalization \citep{ioffe2015batch} was proposed to reduce such undesirable ``covariate shift''. The method normalizes the summed inputs to each hidden unit over the training cases.  Specifically, for the $i^{th}$ summed input in the $l^{th}$ layer, the batch normalization method rescales the summed inputs according to their variances under the distribution of the  data
\begin{equation}
\bar{\act}^l_i = {\gain^l_i\over\sigma^l_i}\left(\act^l_i - \mu^l_i \right)
\qquad
\mu^l_i = \expectation_{\data \sim P(\data)} \left[ \act^l_i \right]
\qquad
\sigma^l_i = \sqrt{\expectation_{\data \sim P(\data)} \left[\left( \act^l_i  - \mu^l_i \right)^2\right]}
\label{eq:bn}
\end{equation}
where $\bar{\act}^l_i$ is normalized summed inputs to the $i^{th}$ hidden unit in the $l^{th}$ layer and $g_i$ is a gain parameter scaling the normalized activation before the non-linear activation function. Note the expectation is under the whole training data distribution.  It is typically impractical to compute the expectations in Eq. (2) exactly, since it would require forward passes through the whole training dataset with the current set of weights.  Instead, $\mu$ and $\sigma$ are estimated using the empirical samples from the current mini-batch. This puts constraints on the size of a mini-batch and it is hard to apply to recurrent neural networks.

\section{Layer normalization}

We now consider the layer normalization method which is designed to overcome the drawbacks of batch normalization. 

Notice that changes in the output of one layer will tend to cause highly correlated changes in the summed inputs to the next layer, especially with ReLU units whose outputs can change by a lot.  
This suggests the ``covariate shift'' problem can be reduced by fixing the mean and the variance of the summed inputs within each layer. We, thus, compute the layer normalization statistics over all the hidden units in the same layer as follows:
\begin{equation}
\mu^l = \frac{1}{H}\sum_{i = 1}^H  \act^l_i
\qquad
\sigma^l = \sqrt{\frac{1}{H}\sum_{i = 1}^H  \left( \act^l_i  - \mu^l \right)^2}
\label{eq:ln}
\end{equation}
where $H$ denotes the number of hidden units in a layer. The difference between Eq. (2) and Eq. (3) is that under layer normalization, all the hidden units in a layer share the same normalization terms $\mu$ and $\sigma$, but different training cases have different normalization terms. 
Unlike batch normalization, layer normaliztion does not impose any constraint on the size of a mini-batch and it can be used in the pure online regime with batch size 1. 

\subsection{Layer normalized recurrent neural networks}

The recent sequence to sequence models \citep{sutskever2014sequence} utilize compact recurrent neural networks to solve sequential prediction problems in natural language processing.  It is common among the NLP tasks to have different sentence lengths for different training cases. This is easy to deal with in an RNN because the same weights are used at every time-step. But when we apply batch normalization to an RNN in the obvious way, we need to to compute and store separate statistics for each time step in a sequence. This is problematic if a test sequence is longer than any of the training sequences. Layer normalization does not have such problem because its normalization terms depend only on the summed inputs to a layer at the current time-step. It also has only one set of gain and bias parameters shared over all time-steps.

In a standard RNN, the summed inputs in the recurrent layer are computed from the current input $\data^t$ and previous vector of hidden states $\Hidden^{t-1}$ which are computed as $\Act^t = \Weights_{hh} \hidden^{t-1} + \Weights_{xh} \data^t$. The layer normalized recurrent layer re-centers and re-scales its activations using the extra normalization terms similar to Eq. (3):
\begin{equation}
\Hidden^t = \actFunc\left[{\Gain\over\sigma^t}\odot\left(\Act^t - \mu^t\right) + \Bias \right]
\qquad
\mu^t = \frac{1}{H}\sum_{i = 1}^H  {a_i^t}
\qquad
\sigma^t = \sqrt{\frac{1}{H}\sum_{i = 1}^H  \left( {a_i^t} - \mu^t \right)^2}
\end{equation}
where $\Weights_{hh}$ is the recurrent hidden to hidden weights and $\Weights_{xh}$ are the bottom up input to hidden weights. $\odot$ is the element-wise multiplication between two vectors. $\Bias$ and $\Gain$ are defined as the bias and gain parameters of the same dimension as $\Hidden^t$.

In a standard RNN, there is a tendency for the average magnitude of the summed inputs to the recurrent units to either grow or shrink at every time-step, leading to exploding or vanishing gradients. In a layer normalized RNN,  the normalization terms make it invariant to re-scaling all of the summed inputs to a layer, which results in much more stable hidden-to-hidden dynamics. 

\section{Related work}
Batch normalization has been previously extended to recurrent neural networks \citep{laurent2015batch, amodei2015deep, cooijmans2016recurrent}. The previous work \citep{cooijmans2016recurrent} suggests the best performance of recurrent batch normalization is obtained by keeping independent normalization statistics for each time-step. The authors show that initializing the gain parameter in the recurrent batch normalization layer to 0.1 makes significant difference in the final performance of the model. Our work is also related to weight normalization \citep{salimans2016weight}.  In weight normalization, instead of the variance, the L2 norm of the incoming weights is used to normalize the summed inputs to a neuron. Applying either weight normalization or batch normalization using expected statistics is equivalent to have a different parameterization of the original feed-forward neural network. Re-parameterization in the ReLU network was studied in the Path-normalized SGD \citep{neyshabur2015path}. Our proposed layer normalization method, however, is not a re-parameterization of the original neural network. The layer normalized model, thus, has different invariance properties than the other methods, that we will study in the following section.

\section{Analysis}
\label{sec:analysis}
In this section, we investigate the invariance properties of different normalization schemes.  

\subsection{Invariance under weights and data transformations}
The proposed layer normalization is related to batch normalization and weight normalization. Although, their normalization scalars are computed differently, these methods can be summarized as normalizing the summed inputs $a_i$ to a neuron through the two scalars $\mu$ and $\sigma$. They also learn an adaptive bias $b$ and gain $g$ for each neuron after the normalization.
\bea
h_i = f({\gain_i\over\sigma_i}\left(a_i - \mu_i \right) + b_i)
\eea
Note that for layer normalization and batch normalization, $\mu$ and $\sigma$ is computed according to Eq. \ref{eq:bn} and \ref{eq:ln}. In weight normalization, $\mu$ is 0, and $\sigma = \|\weights\|_2$. 

Table \ref{tab:inv} highlights the following invariance results for three normalization methods. 

\begin{table}
\scriptsize
\centering
\vspace{-0.2in}
\begin{tabulary}{\linewidth}{L|C|C|C|C|C|C}
\hline
& {Weight matrix}& {Weight matrix} & {Weight vector}  & {Dataset} & {Dataset} & {Single training case}\\
& {re-scaling} & {re-centering} & {re-scaling} & {re-scaling} & {re-centering} & {re-scaling}\\
\hline
\hline
Batch norm & Invariant & No & Invariant & Invariant & Invariant & No\\
Weight norm & Invariant & No & Invariant & No & No & No\\
Layer norm & Invariant & Invariant & No & Invariant & No & Invariant\\
\hline
\end{tabulary}
\caption{Invariance properties under the normalization methods. \vspace{-0.1in}} 
\label{tab:inv}
\vspace{-0.1in}
\end{table}

{\textbf{Weight re-scaling and re-centering:}} First, observe that under batch normalization and weight normalization, any re-scaling to the incoming weights $\weights_i$ of a single neuron has no effect on the normalized summed inputs to a neuron.  To be precise, under batch and weight normalization, if the weight vector is scaled by $\delta$, the two scalar $\mu$ and $\sigma$ will also be scaled by $\delta$. The normalized summed inputs stays the same before and after scaling. So the batch and weight normalization are invariant to the re-scaling of the weights. Layer normalization, on the other hand,  is not invariant to the individual scaling of the single weight vectors. Instead, layer normalization is invariant to scaling of the entire weight matrix and invariant to a shift to all of the incoming weights in the weight matrix. Let there be two sets of model parameters $\theta$, $\theta'$ whose weight matrices $W$ and $W'$ differ by a scaling factor $\delta$ and all of the incoming weights in $W'$ are also shifted by a constant vector $\bm\gamma$, that is $W' = \delta W + {\bf 1}{\bm{\gamma}}^\T$. Under layer normalization, the two models effectively compute the same output:  
\bea
\Hidden' =& f({\Gain\over\sigma'}\left(W'\data - \mu' \right) + \Bias) = f({\Gain\over\sigma'}\left((\delta W + {\bf 1}{\bm{\gamma}}^\T)\data - \mu' \right) + \Bias) \nonumber \\
=& f({\Gain\over\sigma}\left(W\data - \mu \right) + \Bias) = \Hidden.
\eea
Notice that if normalization is only applied to the input before the weights, the model will not be invariant to re-scaling and re-centering of the weights.

{\textbf{Data re-scaling and re-centering:}} We can show that all the normalization methods are invariant to re-scaling the dataset by verifying that the summed inputs of neurons stays constant under the changes. Furthermore, layer normalization is invariant to re-scaling of individual training cases, because the normalization scalars $\mu$ and $\sigma$ in Eq. (3) only depend on the current input data. Let $\data'$ be a new data point obtained by re-scaling $\data$ by $\delta$. Then we have,   
\bea
\hidden_i' =& f({\gain_i\over\sigma'}\left(w_i^\T\data' - \mu' \right) + \bias_i)  
= f({\gain_i\over{\delta\sigma}}\left(\delta w_i^\T\data - \delta\mu \right) + \bias_i) = \hidden_i.
\eea
It is easy to see re-scaling individual data points does not change the model's prediction under layer normalization. Similar to the re-centering of the weight matrix in layer normalization, we can also show that batch normalization is invariant to re-centering of the dataset.

\subsection{Geometry of parameter space during learning}
We have investigated the invariance of the model's prediction under re-centering and re-scaling of the parameters. Learning, however, can behave very differently under different parameterizations, even though the models express the same underlying function. In this section, we analyze learning behavior through the geometry and the manifold of the parameter space. We show that the normalization scalar $\sigma$ can implicitly reduce learning rate and makes learning more stable.

\subsubsection{Riemannian metric}
The learnable parameters in a statistical model form a smooth manifold that consists of all possible input-output relations of the model. For models whose output is a probability distribution, a natural way to measure the separation of two points on this manifold is the Kullback-Leibler divergence between their model output distributions. Under the KL divergence metric, the parameter space is a Riemannian manifold.

The curvature of a Riemannian manifold is entirely captured by its Riemannian metric, whose quadratic form is denoted as $ds^2$. That is the infinitesimal distance in the tangent space at a point in the parameter space. Intuitively, it measures the changes in the model output from the parameter space along a tangent direction. The Riemannian metric under KL was previously studied \citep{amari1998natural} and was shown to be well approximated under second order Taylor expansion using the Fisher information matrix: 
\bea
ds^2 &= \kldiv\big[P(\target \given \data;\  \theta)\| P(\target \given \data ;\  \theta+\delta)\big] \approx \frac{1}{2} \delta^\T \fisher(\theta) \delta, \\
\fisher(\theta) &= \expectation_{\data \sim P(\data), \target \sim P(\target\given\data)}\left[{\partial \log P(\target \given \data ;\ \theta) \over \partial \theta}{\partial \log P(\target \given \data ;\ \theta) \over \partial \theta}^\T\right],
\eea
where, $\delta$ is a small change to the parameters. The Riemannian metric above presents a geometric view of parameter spaces. The following analysis of the Riemannian metric provides some insight into how normalization methods could help in training neural networks.

\subsubsection{The geometry of normalized generalized linear models}
We focus our geometric analysis on the generalized linear model. The results from the following analysis can be easily applied to understand deep neural networks with block-diagonal approximation to the Fisher information matrix, where each block corresponds to the parameters for a single neuron. 

A generalized linear model (GLM) can be regarded as parameterizing an output distribution from the exponential family using a weight vector $w$ and bias scalar $b$. To be consistent with the previous sections, the log likelihood of the GLM can be written using the summed inputs $a$ as the following:
\bea
\log P(y \given \data ;\  w, b) &= \frac{(\act+b) y - \eta(\act+b)}{\phi} + c(y, \phi),\\
\expectation[y \given \data] = f(\act + b) &= f(\weights^\T \data + b), \ \  \text{Var}[\target \given \data] = \phi f'(\act + b),
\eea
where, $f(\cdot)$ is the transfer function that is the analog of the non-linearity in neural networks, $f'(\cdot)$ is the derivative of the transfer function, $\eta(\cdot)$ is a real valued function and $c(\cdot)$ is the log partition function. $\phi$ is a constant that scales the output variance.
Assume a $H$-dimensional output vector $\Target = [y_1, y_2, \cdots, y_H]$ is modeled using $H$ independent GLMs and $\log P(\Target \given \data ;\ W, \Bias) = \sum_{i=1}^H \log P(y_i \given \data ;\ w_i, b_i) $. Let $W$ be the weight matrix whose rows are the weight vectors of the individual GLMs, $\Bias$ denote the bias vector of length $H$ and $\vecop(\cdot)$ denote the Kronecker vector operator.
The Fisher information matrix for the multi-dimensional GLM with respect to its parameters $\theta=[\weights_1^\T, \bias_1,\cdots,\weights_H^\T, \bias_H]^\T = \vecop([W, \Bias]^\T)$ is simply the expected Kronecker product of the data features and the output covariance matrix: 
\bea
\fisher(\theta) &= \expectation_{\data \sim P(\data)}\left[\frac{\text{Cov}[\Target \given \data]}{\phi^2}\otimes\begin{bmatrix}\data\data^\T & \data \\ \data^\T & 1\end{bmatrix}\right].
\eea
We obtain normalized GLMs by applying the normalization methods to the summed inputs $a$ in the original model through $\mu$ and $\sigma$. Without loss of generality, we denote $\bar{\fisher}$ as the Fisher information matrix under the normalized multi-dimensional GLM with the additional gain parameters $\theta=\vecop([W, \Bias, \Gain]^\T)$:
\bea
\bar{\fisher}(\theta) = \begin{bmatrix}[1.5] \bar{\fisher}_{11} & \cdots & \bar{\fisher}_{1H} \\ \vdots & \ddots  & \vdots \\\bar{\fisher}_{H1} &\cdots & \bar{\fisher}_{HH} \end{bmatrix},
\quad
\bar{\fisher}_{ij} &= \expectation_{\data \sim P(\data)}\left[\frac{\text{Cov}[\target_i, \  \target_j \ \given\data]}{\phi^2}\begin{bmatrix}[1.5]\frac{\gain_i\gain_j}{\sigma_i\sigma_j}\normX_i\normX_j^\T & \normX_i\frac{\gain_i}{\sigma_i} & \normX_i\frac{\gain_i(\act_j-\mu_j)}{\sigma_i\sigma_j} \\ \normX_j^\T\frac{\gain_j}{\sigma_j} & 1 & \frac{\act_j-\mu_j}{\sigma_j} \\\normX_j^\T\frac{\gain_j(\act_i-\mu_i)}{\sigma_i\sigma_j} &\frac{\act_i-\mu_i}{\sigma_i} & \frac{(\act_i-\mu_i)(\act_j-\mu_j)}{\sigma_i\sigma_j}\end{bmatrix}\right] \label{eq:nfisher}\\
\normX_i &= \data - \frac{\partial{\mu_i}}{\partial{\weights_i}}  - \frac{\act_i-\mu_i}{\sigma_i}\frac{\partial{\sigma_i}}{\partial{\weights_i}}.
\eea

{\textbf{Implicit learning rate reduction through the growth of the weight vector:}}
Notice that, comparing to standard GLM, the block $\bar{\fisher}_{ij}$ along the weight vector $w_i$ direction is scaled by the gain parameters and the normalization scalar $\sigma_i$. If the norm of the weight vector $w_i$ grows twice as large, even though the model's output remains the same, the Fisher information matrix will be different. The curvature along the $w_i$ direction will change by a factor of $\frac{1}{2}$ because the $\sigma_i$ will also be twice as large. As a result, for the same parameter update in the normalized model, the norm of the weight vector effectively controls the learning rate for the weight vector. During learning, it is harder to change the orientation of the weight vector with large norm.  The normalization methods, therefore, have an implicit ``early stopping'' effect on the weight vectors and help to stabilize learning towards convergence.

{\textbf{Learning the magnitude of incoming weights:}}
In normalized models, the magnitude of the incoming weights is explicitly parameterized by the gain parameters. We compare how the model output changes between updating the gain parameters in the normalized GLM and updating the magnitude of the equivalent weights under original parameterization during learning. The direction along the gain parameters in $\bar{\fisher}$ captures the geometry for the magnitude of the incoming weights.  We show that Riemannian metric along the magnitude of the incoming weights for the standard GLM is scaled by the norm of its input, whereas learning the gain parameters for the batch normalized and layer normalized models depends only on the magnitude of the prediction error.  Learning the magnitude of incoming weights in the normalized model is therefore, more robust to the scaling of the input and its parameters than in the standard model. See Appendix for detailed derivations.

\section{Experimental results}
\label{sec:exp}

We perform experiments with layer normalization on 6 tasks, with a focus on recurrent neural networks: image-sentence ranking, question-answering, contextual language modelling, generative modelling, handwriting sequence generation and MNIST classification. Unless otherwise noted, the default initialization of layer normalization is to set the adaptive gains to $1$ and the biases to $0$ in the experiments.

\subsection{Order embeddings of images and language}

\begin{figure}
  \vspace{-0.3in}
  \centering
  \mbox{
    \subfigure[Recall@1]{\includegraphics[width=0.33\columnwidth]{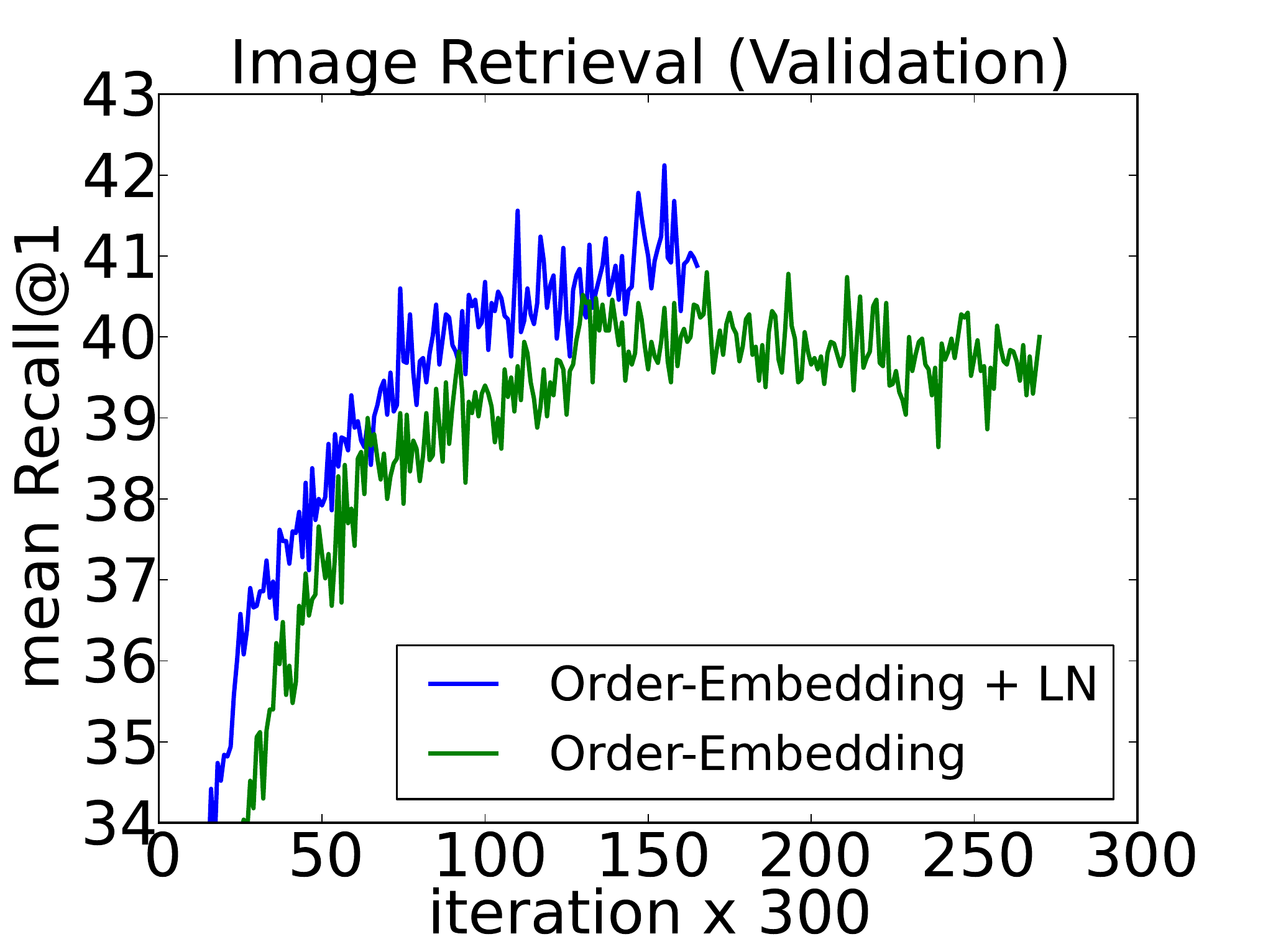}}
    \subfigure[Recall@5]{\includegraphics[width=0.33\columnwidth]{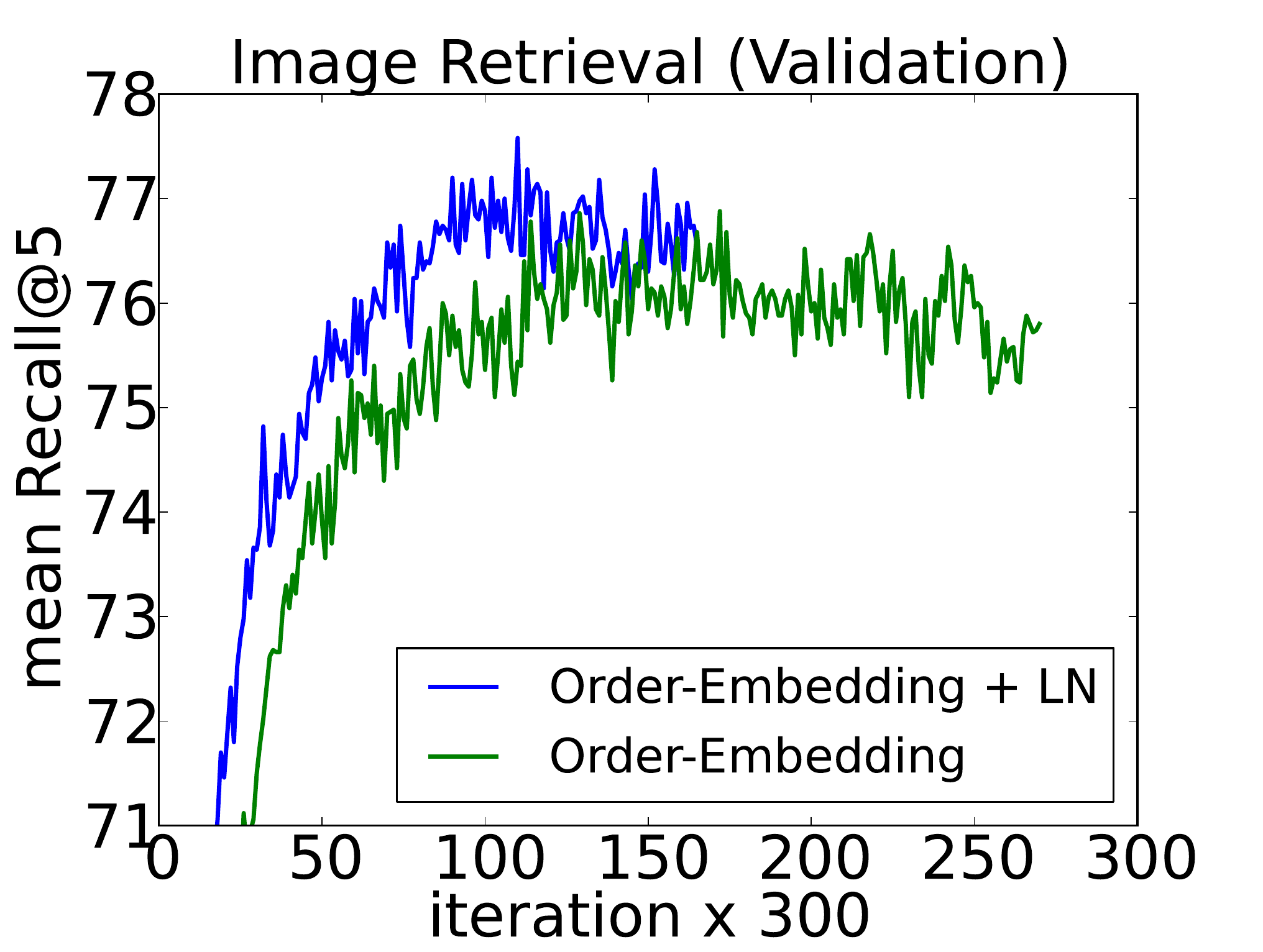}}
    \subfigure[Recall@10]{\includegraphics[width=0.33\columnwidth]{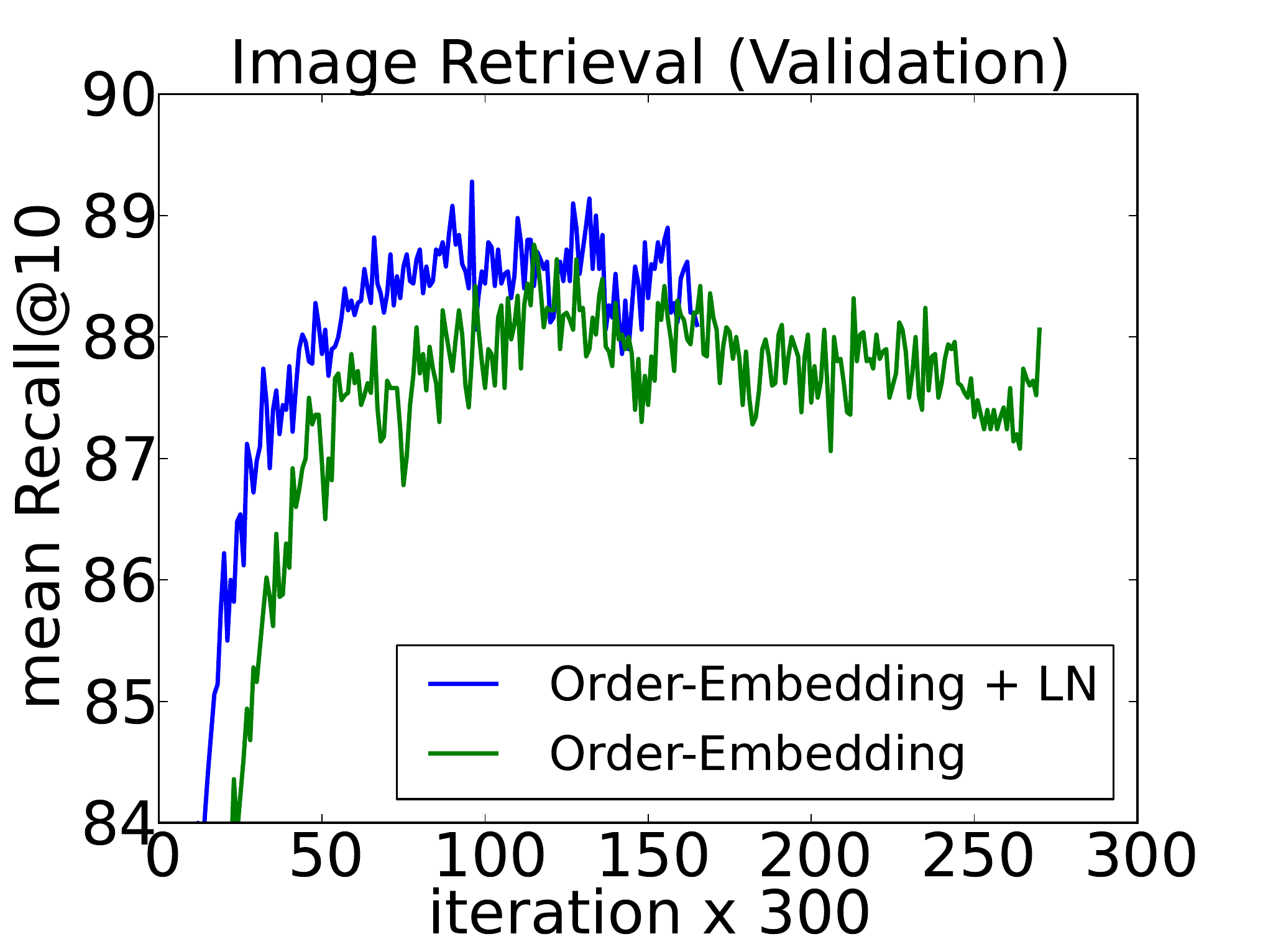}}\quad
  }
 
  \caption{Recall@K curves using order-embeddings with and without layer normalization.
  \vspace{-0.05in}
  }
\label{fig:orderemb}
    ~\\[-.1in]
\end{figure}

\begin{table}
\footnotesize
\centering
\begin{tabulary}{\linewidth}{L|CCCC|CCCC}
\hline
\multicolumn{9}{c}{\textbf{MSCOCO}} \\
\hline
& \multicolumn{4}{c}{Caption Retrieval} & \multicolumn{4}{c}{Image Retrieval} \\
\textbf{Model} & \textbf{R@1} & \textbf{R@5} & \textbf{R@10} & \textbf{Mean} \it{r} & \textbf{R@1} & \textbf{R@5} & \textbf{R@10} & \textbf{Mean} \it{r} \\
\hline
\hline
Sym \citep{vendrov2015order} & 45.4 & & 88.7 & 5.8 & 36.3 & & 85.8 & 9.0 \\
OE \citep{vendrov2015order} & 46.7 & & 88.9 & 5.7 & 37.9 & & 85.9 & 8.1 \\
OE (ours) & 46.6 & 79.3 & 89.1 & 5.2 & 37.8 & 73.6 & 85.7 & 7.9 \\
OE + LN & {\bf 48.5} & {\bf 80.6} & {\bf 89.8} & {\bf 5.1} & {\bf 38.9} & {\bf 74.3} & {\bf 86.3} & {\bf 7.6} \\
\hline
\end{tabulary}
\caption{Average results across 5 test splits for caption and image retrieval. \textbf{R@K} is Recall@K
    (high is good). \textbf{Mean} {\it r} is the mean rank (low is good). Sym corresponds to the symmetric baseline while OE indicates order-embeddings.
  \vspace{-0.15in}
  } 
\label{table:coco}
\end{table}

In this experiment, we apply layer normalization to the recently proposed order-embeddings model of \citet{vendrov2015order} for learning a joint embedding space of images and sentences. We follow the same experimental protocol as \citet{vendrov2015order} and modify their publicly available code to incorporate layer normalization \footnote{\url{https://github.com/ivendrov/order-embedding}} which utilizes Theano \citep{team2016theano}. Images and sentences from the Microsoft COCO dataset \citep{lin2014microsoft} are embedded into a common vector space, where a GRU \citep{cho2014learning} is used to encode sentences and the outputs of a pre-trained VGG ConvNet \citep{simonyan2014very} (10-crop) are used to encode images. The order-embedding model represents images and sentences as a 2-level partial ordering and replaces the cosine similarity scoring function used in \citet{kiros2014unifying} with an asymmetric one.

We trained two models: the baseline order-embedding model as well as the same model with layer normalization applied to the GRU. After every 300 iterations, we compute Recall@K (R@K) values on a held out validation set and save the model whenever R@K improves. The best performing models are then evaluated on 5 separate test sets, each containing 1000 images and 5000 captions, for which the mean results are reported. Both models use Adam \citep{adam} with the same initial hyperparameters and both models are trained using the same architectural choices as used in \citet{vendrov2015order}. We refer the reader to the appendix for a description of how layer normalization is applied to GRU.

Figure $\ref{fig:orderemb}$ illustrates the validation curves of the models, with and without layer normalization. We plot R@1, R@5 and R@10 for the image retrieval task. We observe that layer normalization offers a per-iteration speedup across all metrics and converges to its best validation model in 60\% of the time it takes the baseline model to do so. In Table $\ref{table:coco}$, the test set results are reported from which we observe that layer normalization also results in improved generalization over the original model. The results we report are state-of-the-art for RNN embedding models, with only the structure-preserving model of \citet{wang2015learning} reporting better results on this task. However, they evaluate under different conditions (1 test set instead of the mean over 5) and are thus not directly comparable.

\subsection{Teaching machines to read and comprehend}

\begin{figure}
  \centering
  \vspace{-0.3in}
  \includegraphics[width=1.0\columnwidth]{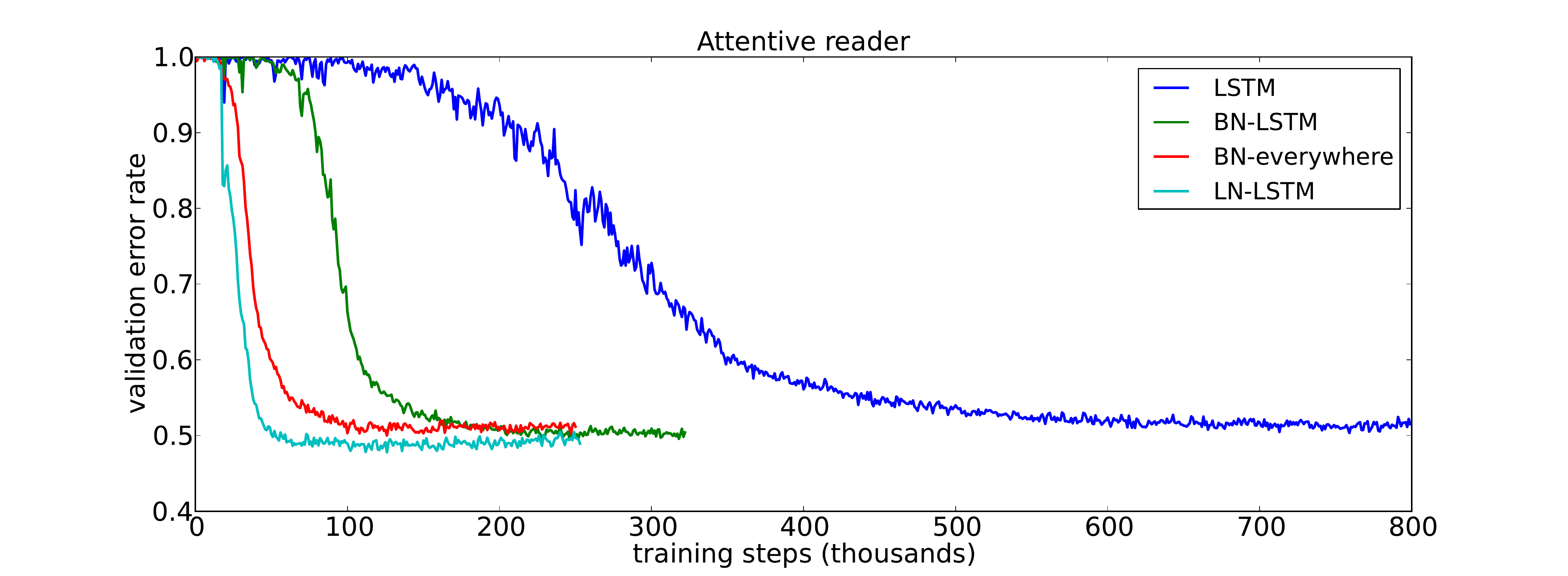}  
  \caption{Validation curves for the attentive reader model. BN results are taken from \citep{cooijmans2016recurrent}.
 \vspace{-0.15in} 
}
  \label{fig:attread}
  \vspace{-0.1in}
\end{figure}

In order to compare layer normalization to the recently proposed recurrent batch normalization \citep{cooijmans2016recurrent}, we train an unidirectional attentive reader model on the CNN corpus both introduced by \citet{hermann2015teaching}. This is a question-answering task where a query description about a passage must be answered by filling in a blank. The data is anonymized such that entities are given randomized tokens to prevent degenerate solutions, which are consistently permuted during training and evaluation. We follow the same experimental protocol as \citet{cooijmans2016recurrent} and modify their public code to incorporate layer normalization \footnote{\url{https://github.com/cooijmanstim/Attentive_reader/tree/bn}} which uses Theano \citep{team2016theano}. We obtained the pre-processed dataset used by \citet{cooijmans2016recurrent} which differs from the original experiments of \citet{hermann2015teaching} in that each passage is limited to 4 sentences. In \citet{cooijmans2016recurrent}, two variants of recurrent batch normalization are used: one where BN is only applied to the LSTM while the other applies BN everywhere throughout the model. In our experiment, we only apply layer normalization within the LSTM.

The results of this experiment are shown in Figure $\ref{fig:attread}$. We observe that layer normalization not only trains faster but converges to a better validation result over both the baseline and BN variants. In \citet{cooijmans2016recurrent}, it is argued that the scale parameter in BN must be carefully chosen and is set to 0.1 in their experiments. We experimented with layer normalization for both 1.0 and 0.1 scale initialization and found that the former model performed significantly better. This demonstrates that layer normalization is not sensitive to the initial scale in the same way that recurrent BN is. \footnote{We only produce results on the validation set, as in the case of \citet{cooijmans2016recurrent}}

\subsection{Skip-thought vectors}
\begin{figure}
  \centering
  \vspace{-0.5in}
  \mbox{
    \subfigure[SICK($r$)]{\includegraphics[width=0.33\columnwidth]{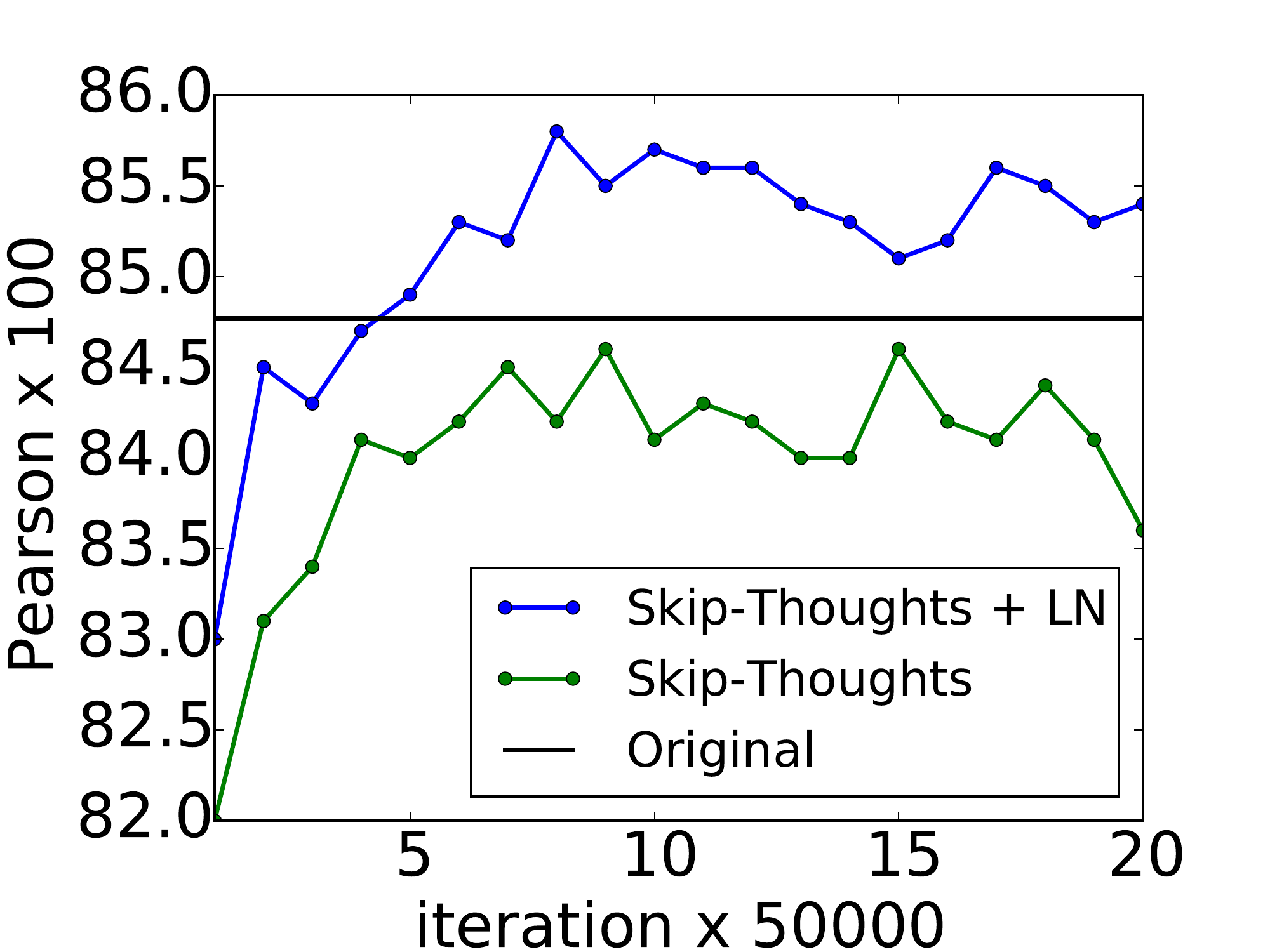}}
    \subfigure[SICK(MSE)]{\includegraphics[width=0.33\columnwidth]{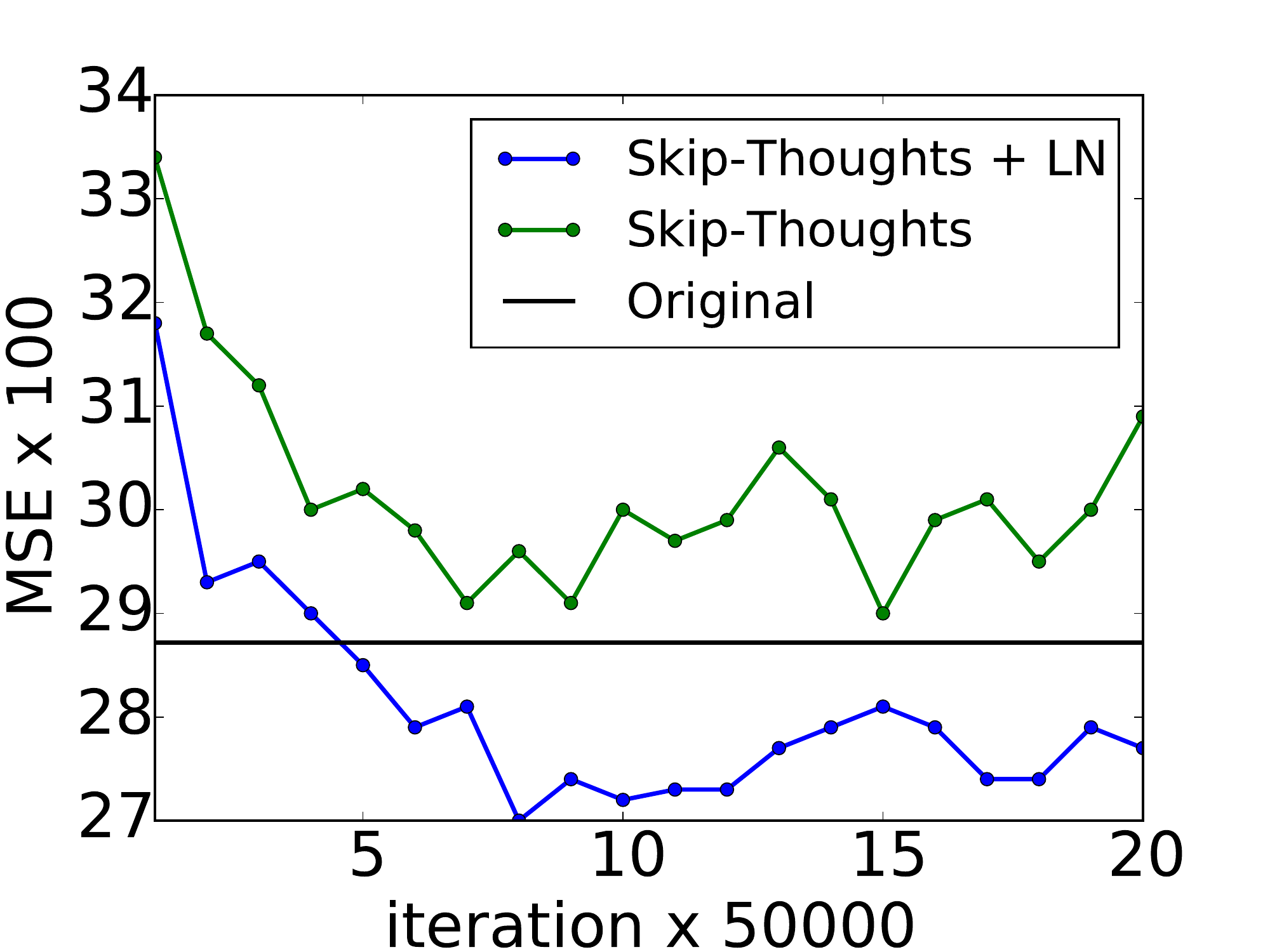}}
    \subfigure[MR]{\includegraphics[width=0.33\columnwidth]{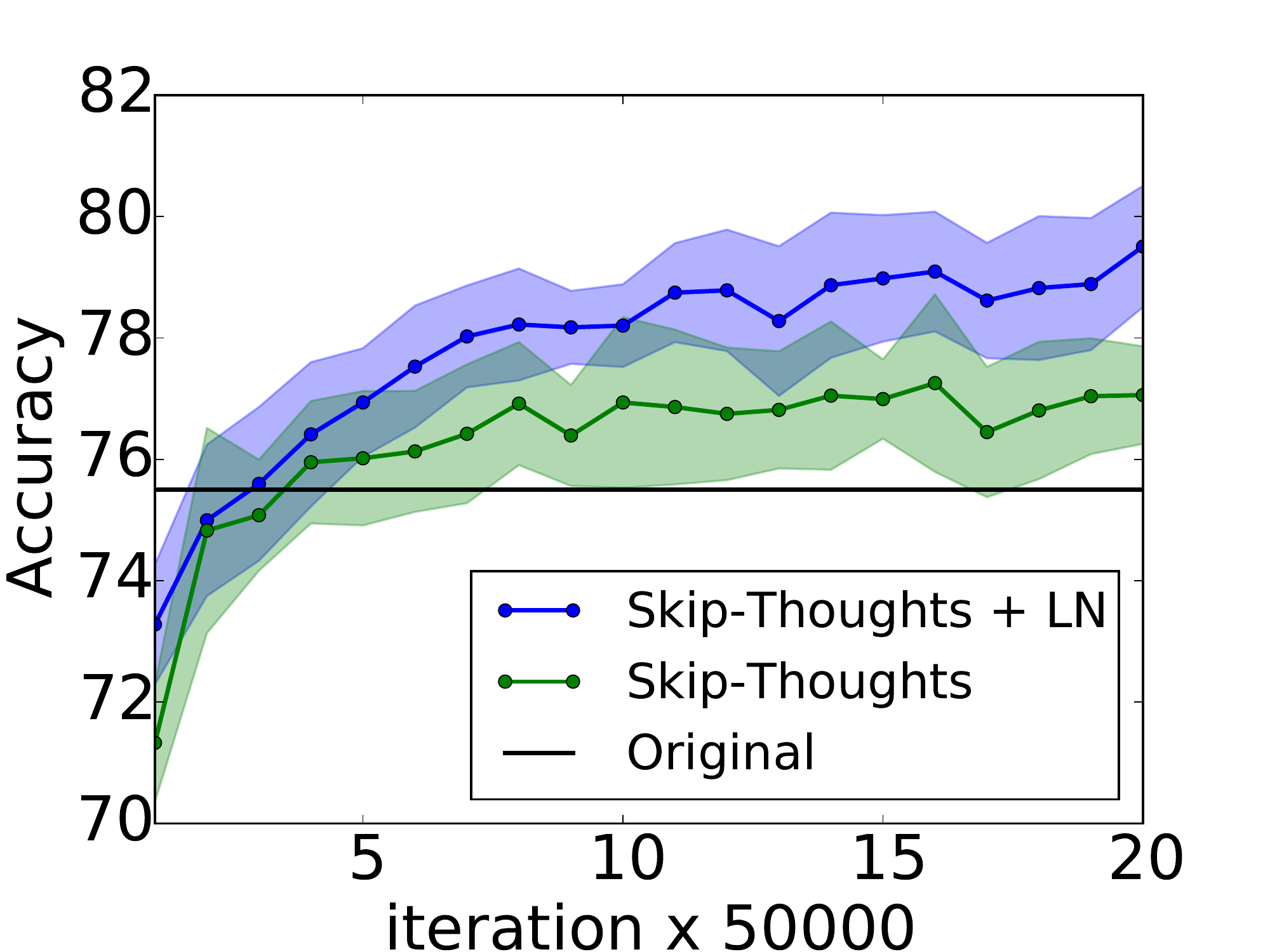}}\quad
  }
    ~\\[-.15in]
    \mbox{
    \subfigure[CR]{\includegraphics[width=0.33\columnwidth]{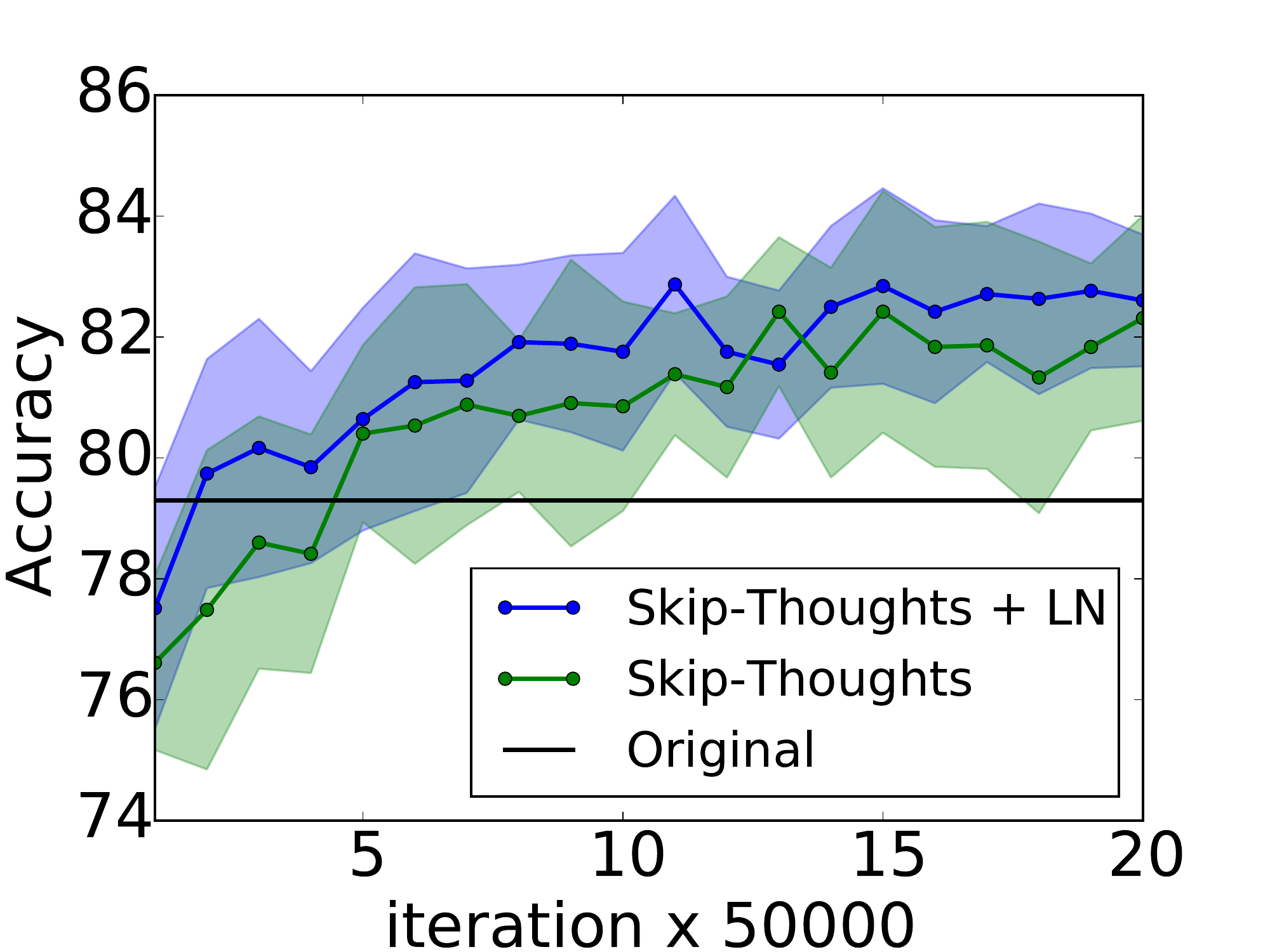}}
    \subfigure[SUBJ]{\includegraphics[width=0.33\columnwidth]{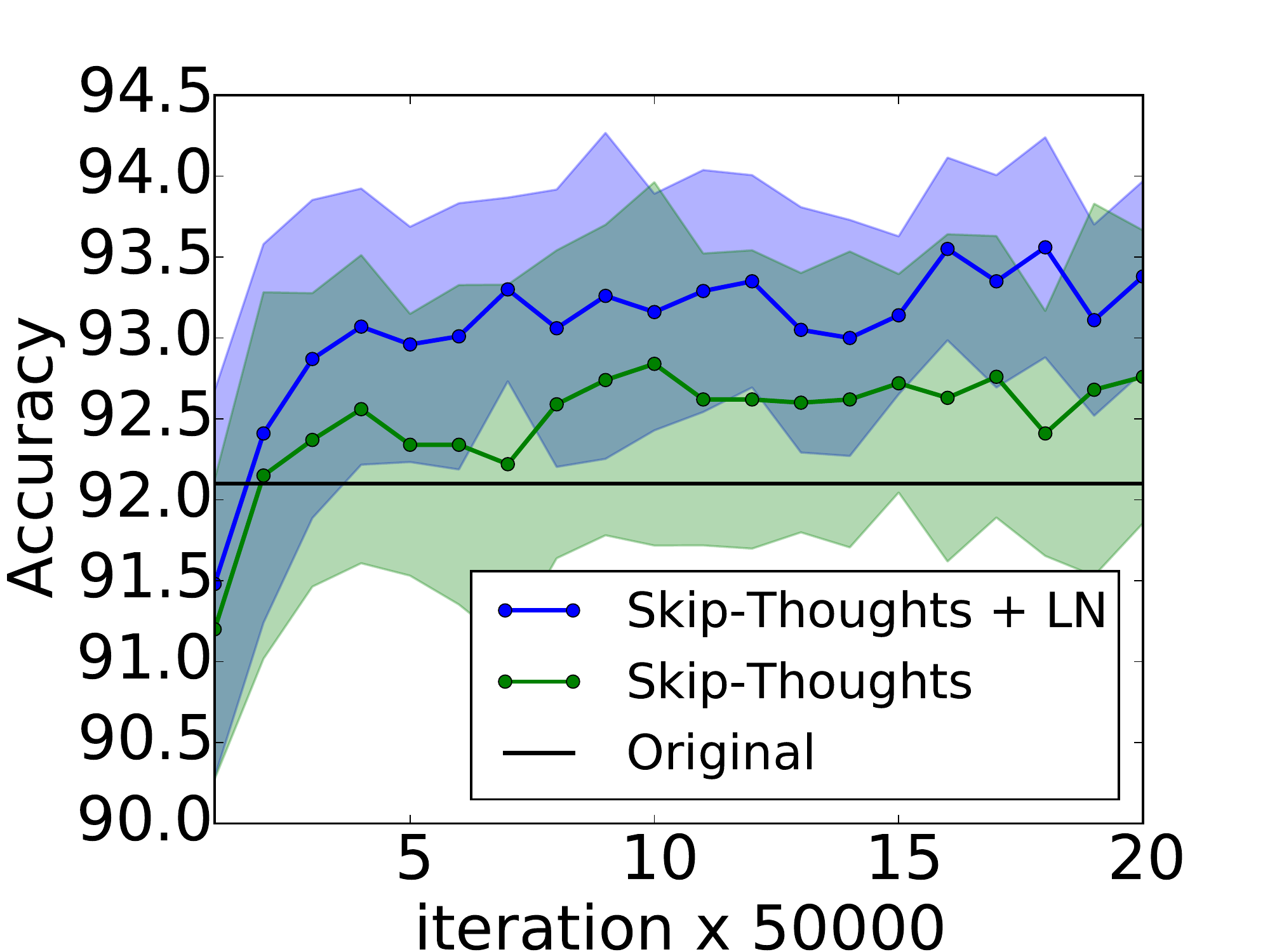}}
    \subfigure[MPQA]{\includegraphics[width=0.33\columnwidth]{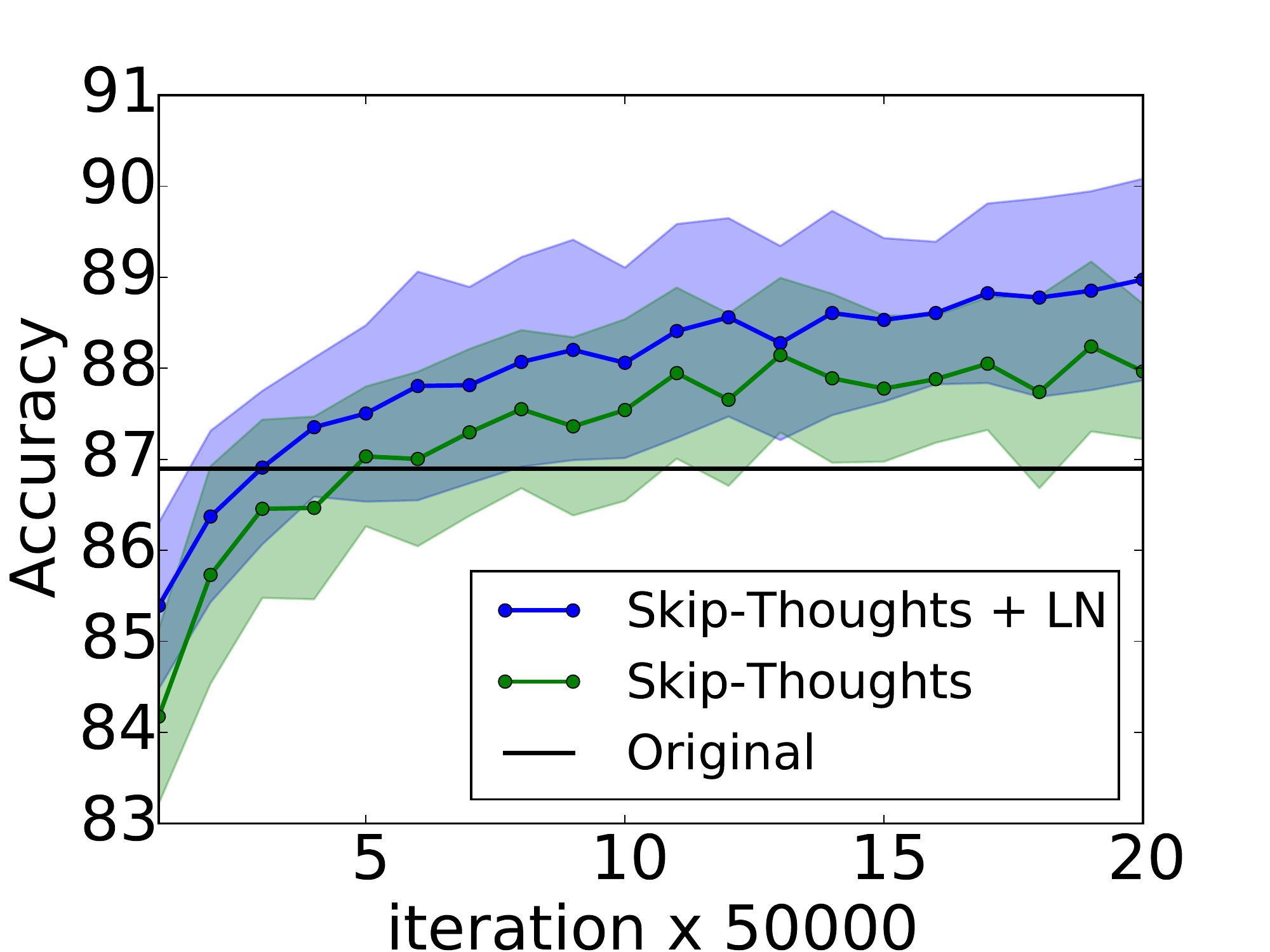}}\quad
  }
 
  \caption{Performance of skip-thought vectors with and without layer normalization on downstream tasks as a function of training iterations. The original lines are the reported results in \citep{kiros2015skip}. Plots with error use 10-fold cross validation. Best seen in color.}
\label{fig:skipthoughts}
\end{figure}

Skip-thoughts \citep{kiros2015skip} is a generalization of the skip-gram model \citep{mikolov2013efficient} for learning unsupervised distributed sentence representations. Given contiguous text, a sentence is encoded with a encoder RNN and decoder RNNs are used to predict the surrounding sentences. \citet{kiros2015skip} showed that this model could produce generic sentence representations that perform well on several tasks without being fine-tuned. However, training this model is time-consuming, requiring several days of training in order to produce meaningful results.

\begin{table}
\small
\centering
\begin{tabular}{lccccccc}
\toprule \bf Method & \bf SICK($r$) & \bf SICK($\rho$) & {\bf SICK(}MSE{\bf)} & \bf MR & \bf CR & \bf SUBJ & \bf MPQA \\ \midrule
Original \citep{kiros2015skip} & 0.848 & 0.778 & 0.287 & 75.5 & 79.3 & 92.1 & 86.9  \\ \midrule
Ours & 0.842 & 0.767 & 0.298 & 77.3 & 81.8 & 92.6 & 87.9  \\
Ours + LN & 0.854 & 0.785 & 0.277 & {\bf 79.5} & 82.6 & 93.4 & 89.0   \\
Ours + LN $\dagger$ & {\bf 0.858} & {\bf 0.788} & {\bf 0.270} & 79.4 & {\bf 83.1} & {\bf 93.7} & {\bf 89.3}   \\ \bottomrule
\end{tabular}
    ~\\[-.05in]
\caption{Skip-thoughts results. The first two evaluation columns indicate Pearson and Spearman correlation, the third is mean squared error and the remaining indicate classification accuracy. Higher is better for all evaluations except MSE. Our models were trained for 1M iterations with the exception of ($\dagger$) which was trained for 1 month (approximately 1.7M iterations)}
\label{tab:skipthoughts}
\end{table}

In this experiment we determine to what effect layer normalization can speed up training. Using the publicly available code of \citet{kiros2015skip} \footnote{\url{https://github.com/ryankiros/skip-thoughts}}, we train two models on the BookCorpus dataset \citep{zhu2015aligning}: one with and one without layer normalization. These experiments are performed with Theano \citep{team2016theano}. We adhere to the experimental setup used in \citet{kiros2015skip}, training a 2400-dimensional sentence encoder with the same hyperparameters. Given the size of the states used, it is conceivable layer normalization would produce slower per-iteration updates than without. However, we found that provided CNMeM \footnote{\url{https://github.com/NVIDIA/cnmem}} is used, there was no significant difference between the two models. We checkpoint both models after every 50,000 iterations and evaluate their performance on five tasks: semantic-relatedness (SICK) \citep{marelli2014semeval}, movie review sentiment (MR) \citep{pang2005seeing}, customer product reviews (CR) \citep{hu2004mining}, subjectivity/objectivity classification (SUBJ) \citep{pang2004sentimental} and opinion polarity (MPQA) \citep{wiebe2005annotating}. We plot the performance of both models for each checkpoint on all tasks to determine whether the performance rate can be improved with LN.

The experimental results are illustrated in Figure $~\ref{fig:skipthoughts}$. We observe that applying layer normalization results both in speedup over the baseline as well as better final results after 1M iterations are performed as shown in Table $~\ref{tab:skipthoughts}$. We also let the model with layer normalization train for a total of a month, resulting in further performance gains across all but one task. We note that the performance differences between the original reported results and ours are likely due to the fact that the publicly available code does not condition at each timestep of the decoder, where the original model does.

\subsection{Modeling binarized MNIST using DRAW}
\begin{wrapfigure}{r}{5cm}
  \vspace{-0.25in}
  \centering
\includegraphics[width=1.0\columnwidth]{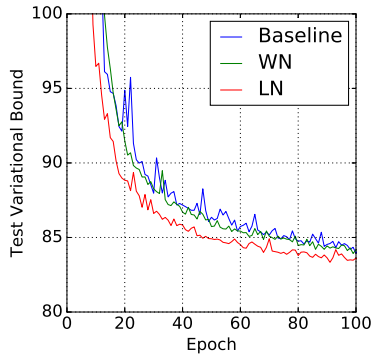}  
\caption{DRAW model test negative log likelihood with and without layer normalization.
\vspace{-0.15in}
}
\label{fig:draw}
\end{wrapfigure}

We also experimented with the generative modeling on the MNIST
dataset.  Deep Recurrent Attention Writer (DRAW) \citep{draw} has
previously achieved the state-of-the-art performance on modeling the
distribution of MNIST digits. The model uses a differential attention
mechanism and a recurrent neural network to sequentially generate
pieces of an image. We evaluate the effect of layer normalization
on a DRAW model using 64 glimpses and 256 LSTM hidden units. The
model is trained with the default setting of Adam \citep{adam}
optimizer and the minibatch size of 128. Previous publications on
binarized MNIST have used various training protocols to generate
their datasets. In this experiment, we used the fixed binarization
from \citet{hugobmnist}. The dataset has been split into 50,000
training, 10,000 validation and 10,000 test images.

Figure \ref{fig:draw} shows the test variational bound for the first 100 epoch. It highlights the speedup benefit of applying layer normalization that the layer normalized DRAW converges
almost twice as fast than the baseline model. After 200 epoches, the baseline model converges
to a variational log likelihood of 82.36 nats  on the test data and
the layer normalization model obtains 82.09 nats.

\subsection{Handwriting sequence generation}

\begin{figure}
  \vspace{-0.3in}
  \centering
\includegraphics[width=0.9\columnwidth]{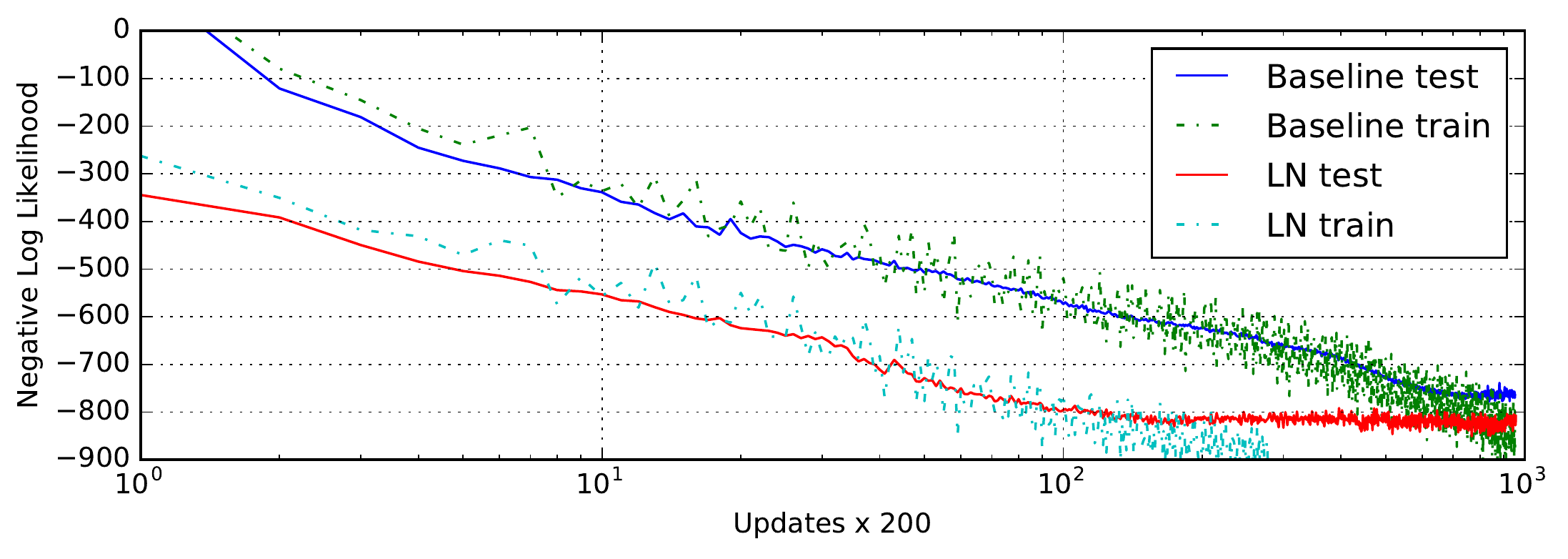}  
\caption{Handwriting sequence generation model negative log likelihood with and without layer normalization. The models are trained with mini-batch size of 8 and sequence length of 500,
\vspace{-0.15in}
}
\label{fig:seqgen}
\vspace{-0.05in}
\end{figure}
The previous experiments mostly examine RNNs on NLP tasks whose lengths are in the range of 10 to 40. To show the effectiveness of layer normalization on longer sequences, we performed handwriting generation tasks using the IAM Online Handwriting Database \citep{liwicki2005iam}. IAM-OnDB consists of handwritten lines collected from 221 different writers. When given the input character string, the goal is to predict a sequence of x and y pen co-ordinates of the corresponding handwriting line on the whiteboard. There are, in total, 12179 handwriting line sequences. The input string is typically more than 25 characters and the average handwriting line has a length around 700.  

We used the same model architecture as in Section (5.2) of \citet{graves2013generating}. The model architecture consists of three hidden layers
of 400 LSTM cells, which produce 20 bivariate Gaussian mixture components at the output layer, and a size 3 input layer. The character sequence was encoded with one-hot vectors, and hence the window vectors were size 57. A mixture of 10 Gaussian functions was used for the window parameters, requiring a size 30 parameter vector. The total number of weights was increased to approximately 3.7M. The model is trained using mini-batches of size 8 and the Adam \citep{adam} optimizer.

The combination of small mini-batch size and very long sequences makes it important to have very stable hidden dynamics. Figure \ref{fig:seqgen} shows that layer normalization converges to a comparable log likelihood as the baseline model but is much faster.

\subsection{Permutation invariant MNIST}

In addition to RNNs, we investigated layer normalization in feed-forward networks. We show how layer normalization compares with batch normalization on the well-studied permutation invariant MNIST classification problem. From the previous analysis, layer normalization is invariant to input re-scaling which is desirable for the internal hidden layers. But this is unnecessary for the logit outputs where the prediction confidence is determined by the scale of the logits. We only apply layer normalization to the fully-connected hidden layers that excludes the last softmax layer.  

\begin{figure}
  \vspace{-0.6in}
  \centering
\includegraphics[width=0.9\columnwidth]{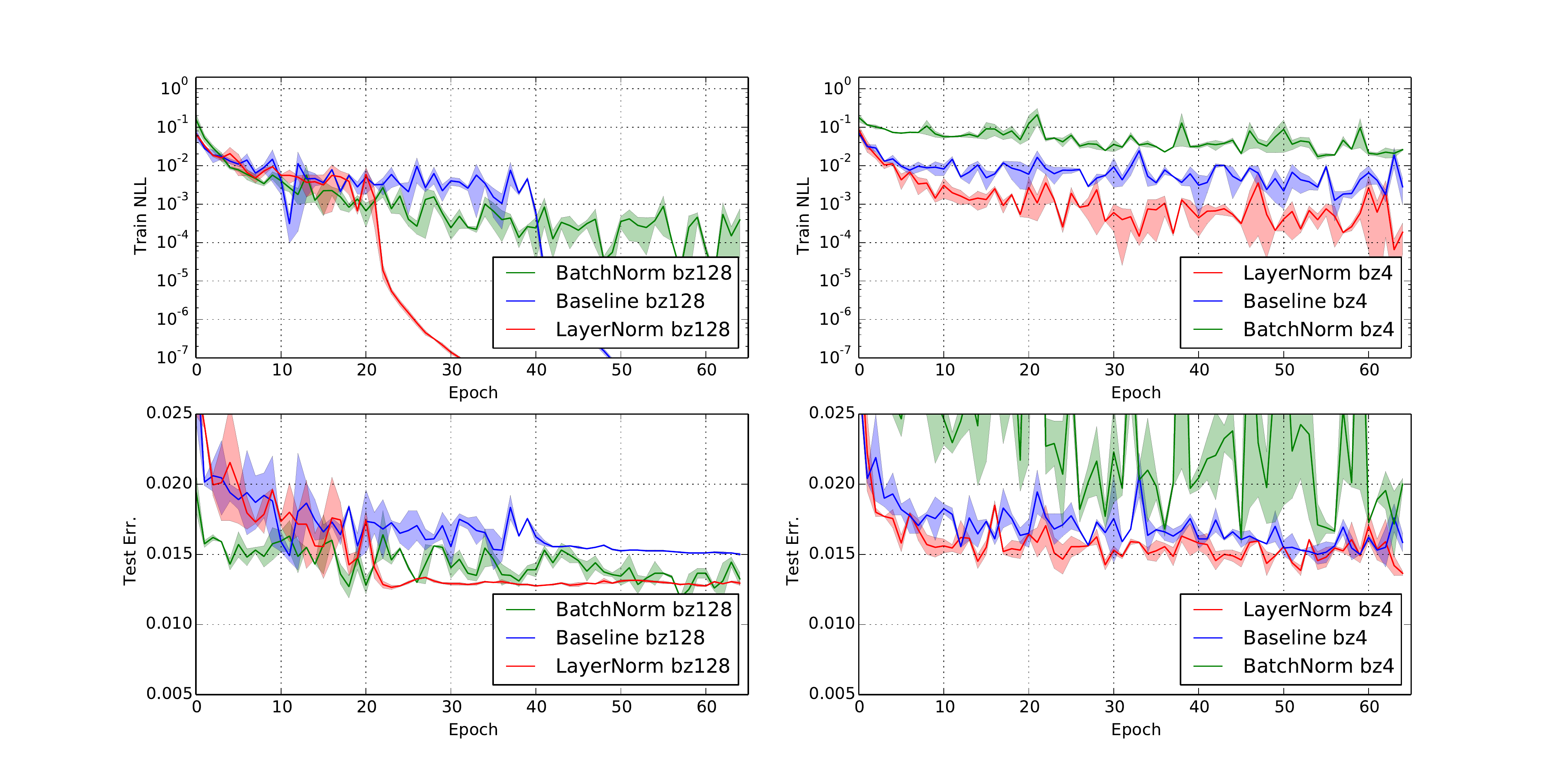}  
\caption{Permutation invariant MNIST 784-1000-1000-10 model negative log likelihood and test error with layer normalization and batch normalization. (Left) The models are trained with batch-size of 128. (Right) The models are trained with batch-size of 4. 
\vspace{-0.15in}
}
\label{fig:mnist}
\vspace{-0.05in}
\end{figure}

All the models were trained using 55000 training data points and the Adam \citep{adam} optimizer. For the smaller batch-size, the variance term for batch normalization is computed using the unbiased estimator.
The experimental results from Figure \ref{fig:mnist} highlight that layer normalization is robust to the batch-sizes and exhibits a faster training convergence comparing to batch normalization that is applied to all layers.

\subsection{Convolutional Networks}
\label{sec:convnets}

We have also experimented with convolutional neural networks. In our preliminary experiments, we observed that layer normalization offers a speedup over the baseline model without normalization, but batch normalization outperforms the other methods. With fully connected layers, all the hidden units in a layer tend to make similar contributions to the final prediction and re-centering and re-scaling the summed inputs to a layer works well.  However, the assumption of similar contributions is no longer true for convolutional neural networks. The large number of the hidden units whose receptive fields lie near the boundary of the image are rarely turned on and thus have very different statistics from the rest of the hidden units within the same layer. We think further research is needed to make layer normalization work well in ConvNets.

\section{Conclusion}
\label{sec:discussion}

In this paper, we introduced layer normalization to speed-up the  training of neural networks. We provided a theoretical analysis that compared the invariance properties of layer normalization with batch normalization and weight normalization. We showed that layer normalization is invariant to per training-case feature shifting and scaling.

Empirically, we showed that recurrent neural networks benefit the most from the proposed method especially for long sequences and small mini-batches.

\section*{Acknowledgments}   
\vspace{-0.1in}
This research was funded by grants from NSERC, CFI, and Google.

\pagebreak
\setlength{\bibsep}{5pt plus 0.3ex}
\begin{small}
\vspace{-0.1in}
\bibliography{ln_biblio}
\end{small}

\newpage

\section*{Supplementary Material}

\subsection*{Application of layer normalization to each experiment}

This section describes how layer normalization is applied to each of the papers' experiments. For notation convenience, we define layer normalization as a function mapping $LN: \real^D \to \real^D$ with two set of adaptive parameters, gains $\bm{\alpha}$ and biases $\bm{\beta}$:
\begin{eqnarray}
LN({\bf z} ; \bm{\alpha}, \bm{\beta}) = \frac{({\bf z} - \mu)}{\sigma} \odot \bm{\alpha} + \bm{\beta}, \\
\mu = \frac{1}{D}\sum_{i=1}^D z_i, \quad \sigma = \sqrt{\frac{1}{D}\sum_{i=1}^D (z_i-\mu)^2}, 
\end{eqnarray}
where, $z_i$ is the $i^{th}$ element of the vector ${\bf z}$.

\subsubsection*{Teaching machines to read and comprehend and handwriting sequence generation}

The basic LSTM equations used for these experiment are given by:
\begin{eqnarray}
\begin{pmatrix}{\bf f}_t\\{\bf i}_t\\{\bf o}_t\\{\bf g}_t\end{pmatrix} &=& {\bf W}_h {\bf h}_{t-1} + {\bf W}_x {\bf x}_t + b \\
{\bf c}_t &=& \sigma({\bf f}_t) \odot {\bf c}_{t-1} + \sigma({\bf i}_t) \odot \text{tanh}({\bf g}_t) \\
{\bf h}_t &=& \sigma({\bf o}_t) \odot \text{tanh}({\bf c}_t)
\end{eqnarray}
The version that incorporates layer normalization is modified as follows:
\begin{eqnarray}
\begin{pmatrix}{\bf f}_t\\{\bf i}_t\\{\bf o}_t\\{\bf g}_t\end{pmatrix} &=& LN({\bf W}_h {\bf h}_{t-1}; \bm{\alpha}_1, \bm{\beta}_1) + LN({\bf W}_x {\bf x}_t; \bm{\alpha}_2, \bm{\beta}_2) + b \\
{\bf c}_t &=& \sigma({\bf f}_t) \odot {\bf c}_{t-1} + \sigma({\bf i}_t) \odot \text{tanh}({\bf g}_t) \\
{\bf h}_t &=& \sigma({\bf o}_t) \odot \text{tanh}(LN({\bf c}_t; \bm{\alpha}_3, \bm{\beta}_3))
\end{eqnarray}
where $\bm{\alpha}_i, \bm{\beta}_i$ are the additive and multiplicative parameters, respectively. Each $\bm{\alpha}_i$ is initialized to a vector of zeros and each $\bm{\beta}_i$ is initialized to a vector of ones.

\subsubsection*{Order embeddings and skip-thoughts}

These experiments utilize a variant of gated recurrent unit which is defined as follows:
\begin{eqnarray}
\begin{pmatrix}{\bf z}_t\\{\bf r}_t\end{pmatrix} &=& {\bf W}_h {\bf h}_{t-1} + {\bf W}_x {\bf x}_t \\
\hat{{\bf h}}_t &=& \text{tanh}({\bf W} {\bf x}_t + \sigma({\bf r}_t) \odot ({\bf U} {\bf h}_{t-1})) \\
{\bf h}_t &=& (1 - \sigma({\bf z}_t)) {\bf h}_{t-1} + \sigma({\bf z}_t) \hat{{\bf h}}_t
\end{eqnarray}
Layer normalization is applied as follows:
\begin{eqnarray}
\begin{pmatrix}{\bf z}_t\\{\bf r}_t\end{pmatrix} &=& LN({\bf W}_h {\bf h}_{t-1}; \bm{\alpha}_1, \bm{\beta}_1) + LN({\bf W}_x {\bf x}_t; \bm{\alpha}_2, \bm{\beta}_2) \\
\hat{{\bf h}}_t &=& \text{tanh}(LN({\bf W} {\bf x}_t; \bm{\alpha}_3, \bm{\beta}_3) + \sigma({\bf r}_t) \odot LN({\bf U} {\bf h}_{t-1}; \bm{\alpha}_4, \bm{\beta}_4)) \\
{\bf h}_t &=& (1 - \sigma({\bf z}_t)) {\bf h}_{t-1} + \sigma({\bf z}_t) \hat{{\bf h}}_t
\end{eqnarray}
just as before, $\bm{\alpha}_i$ is initialized to a vector of zeros and each $\bm{\beta}_i$ is initialized to a vector of ones.

\subsubsection*{Modeling binarized MNIST using DRAW}

The layer norm is only applied to the output of the LSTM hidden states in this experiment:

The version that incorporates layer normalization is modified as follows:
\begin{eqnarray}
\begin{pmatrix}{\bf f}_t\\{\bf i}_t\\{\bf o}_t\\{\bf g}_t\end{pmatrix} &=& {\bf W}_h {\bf h}_{t-1} + {\bf W}_x {\bf x}_t + b \\
{\bf c}_t &=& \sigma({\bf f}_t) \odot {\bf c}_{t-1} + \sigma({\bf i}_t) \odot \text{tanh}({\bf g}_t) \\
{\bf h}_t &=& \sigma({\bf o}_t) \odot \text{tanh}(LN({\bf c}_t; \bm{\alpha}, \bm{\beta}))
\end{eqnarray}
where $\bm{\alpha}, \bm{\beta}$ are the additive and multiplicative parameters, respectively. $\bm{\alpha}$ is initialized to a vector of zeros and $\bm{\beta}$ is initialized to a vector of ones.
\subsubsection*{Learning the magnitude of incoming weights}
We now compare how gradient descent updates changing magnitude of the equivalent weights between the normalized GLM and original parameterization. The magnitude of the weights are explicitly parameterized using the gain parameter in the normalized model. Assume there is a gradient update that changes norm of the weight vectors by $\delta_g$. We can project the gradient updates to the weight vector for the normal GLM. The KL metric, ie how much the gradient update changes the model prediction,  for the normalized model depends only on the magnitude of the prediction error. Specifically,

under batch normalization:
\bea
ds^2 = \frac{1}{2}\vecop([0, 0, \delta_g]^\T)^\T \bar{F}(\vecop([W, \Bias, \Gain]^\T) \vecop([0, 0, \delta_g]^\T) = \frac{1}{2}\delta_g^\T \expectation_{\data \sim P(\data)}\left[\frac{\text{Cov}[\Target \given \data]}{\phi^2}\right] \delta_g .
\eea

Under layer normalization:
\bea
ds^2 =& \frac{1}{2}\vecop([0, 0, \delta_g]^\T)^\T \bar{F}(\vecop([W, \Bias, \Gain]^\T) \vecop([0, 0, \delta_g]^\T) \nonumber \\
=& \frac{1}{2}\delta_g^\T  \frac{1}{{\phi^2}}\expectation_{\data \sim P(\data)}\begin{bmatrix}[1.5] {\text{Cov}(y_1, y_1 \given \data)}\frac{(a_1-\mu)^2}{\sigma^2} & \cdots & {\text{Cov}(y_1, y_H \given \data)}\frac{(a_1-\mu)(a_H-\mu)}{\sigma^2} \\ \vdots & \ddots  & \vdots \\\text{Cov}(y_H, y_1 \given \data)\frac{(a_H-\mu)(a_1-\mu)}{\sigma^2} &\cdots & \text{Cov}(y_H, y_H \given \data)\frac{(a_H-\mu)^2}{\sigma^2} \end{bmatrix}   \delta_g 
\eea

Under weight normalization:
\bea
ds^2 =& \frac{1}{2}\vecop([0, 0, \delta_g]^\T)^\T \bar{F}(\vecop([W, \Bias, \Gain]^\T) \vecop([0, 0, \delta_g]^\T) \nonumber \\
=& \frac{1}{2}\delta_g^\T  \frac{1}{{\phi^2}}\expectation_{\data \sim P(\data)}\begin{bmatrix}[1.5] {\text{Cov}(y_1, y_1 \given \data)}\frac{a_1^2}{\|w_1\|_2^2} & \cdots & {\text{Cov}(y_1, y_H \given \data)}\frac{a_1a_H}{\|w_1\|_2\|w_H\|_2} \\ \vdots & \ddots  & \vdots \\\text{Cov}(y_H, y_1 \given \data)\frac{a_Ha_1}{\|w_H\|_2\|w_1\|_2} &\cdots & \text{Cov}(y_H, y_H \given \data)\frac{a_H^2}{\|w_H\|^2_2} \end{bmatrix}   \delta_g.
\eea
Whereas, the KL metric in the standard GLM is related to its activities $a_i = w_i^\T \data$, that is depended on both its current weights and input data. We project the gradient updates to the gain parameter $\delta_gi$ of the $i^{th}$ neuron to its weight vector as $\delta_{gi} \frac{w_i}{\|w_i\|_2}$ in the standard GLM model:
\bea
&\frac{1}{2}\vecop([\delta_{gi} \frac{w_i}{\|w_i\|_2}, 0, \delta_{gj} \frac{w_j}{\|w_i\|_2}, 0]^\T)^\T {F}([w_i^\T, b_i, w_j^\T, b_j]^\T) \vecop([\delta_{gi} \frac{w_i}{\|w_i\|_2}, 0, \delta_{gj} \frac{w_j}{\|w_j\|_2}, 0]^\T) \nonumber \\=& \frac{\delta_{gi}\delta_{gj}}{2\phi^2}  \expectation_{\data \sim P(\data)}\left[{\text{Cov}(y_i, y_j \given \data)}\frac{a_ia_j}{\|w_i\|_2\|w_j\|_2}\right] 
\eea

The batch normalized and layer normalized models are therefore more robust to the scaling of the input and its parameters than the standard model.

\end{document}

%% file: macros.tex
\newcommand{\cut}[1]{}

\newcommand{\T}{\top}
\newcommand{\real}{\mathbb{R}}

\newcommand{\be}{\begin{equation}}
\newcommand{\ee}{\end{equation}}
\def\bea#1\eea{\begin{align}#1\end{align}}
\def\bean#1\eean{\begin{align*}#1\end{align*}}
\makeatletter
\renewcommand*\env@matrix[1][\arraystretch]{%
  \edef\arraystretch{#1}%
  \hskip -\arraycolsep
  \let\@ifnextchar\new@ifnextchar
  \array{*\c@MaxMatrixCols c}}
\makeatother

\DeclareMathOperator{\vecop}{vec}
\newcommand{\Target}{{\bold y}}
\newcommand{\normX}{{\chi}}
\newcommand{\noise}{{\xi}}
\newcommand{\actFunc}{{f}}
\newcommand{\gain}{{g}}
\newcommand{\Gain}{{\bold g}}
\newcommand{\bias}{{b}}
\newcommand{\Bias}{{\bold b}}
\newcommand{\Weights}{{W}}
\newcommand{\weights}{{w}}
\newcommand{\hidden}{{h}}
\newcommand{\Hidden}{{\bold h}}
\newcommand{\act}{{a}}
\newcommand{\Act}{{\bold a}}
\newcommand{\fisher}{{F}}
\newcommand{\data}{{\bold x}}

\newcommand{\expectation}{\mathop{\mathbb{E}}}
\newcommand{\kldiv}{\mathrm{D}_{\rm KL}}

\newcommand{\given}{\,|\,}

\newcommand{\target}{y}

%% file: ln_paper_arxiv.bbl
\begin{thebibliography}{32}
\providecommand{\natexlab}[1]{#1}
\providecommand{\url}[1]{\texttt{#1}}
\expandafter\ifx\csname urlstyle\endcsname\relax
  \providecommand{\doi}[1]{doi: #1}\else
  \providecommand{\doi}{doi: \begingroup \urlstyle{rm}\Url}\fi

\bibitem[Krizhevsky et~al.(2012)Krizhevsky, Sutskever, and
  Hinton]{krizhevsky2012imagenet}
Alex Krizhevsky, Ilya Sutskever, and Geoffrey~E Hinton.
\newblock Imagenet classification with deep convolutional neural networks.
\newblock In \emph{NIPS}, 2012.

\bibitem[Hinton et~al.(2012)Hinton, Deng, Yu, Dahl, Mohamed, Jaitly, Senior,
  Vanhoucke, Nguyen, Sainath, et~al.]{hinton2012deep}
Geoffrey Hinton, Li~Deng, Dong Yu, George~E Dahl, Abdel-rahman Mohamed, Navdeep
  Jaitly, Andrew Senior, Vincent Vanhoucke, Patrick Nguyen, Tara~N Sainath,
  et~al.
\newblock Deep neural networks for acoustic modeling in speech recognition: The
  shared views of four research groups.
\newblock \emph{IEEE}, 2012.

\bibitem[Dean et~al.(2012)Dean, Corrado, Monga, Chen, Devin, Mao, Senior,
  Tucker, Yang, Le, et~al.]{dean2012large}
Jeffrey Dean, Greg Corrado, Rajat Monga, Kai Chen, Matthieu Devin, Mark Mao,
  Andrew Senior, Paul Tucker, Ke~Yang, Quoc~V Le, et~al.
\newblock Large scale distributed deep networks.
\newblock In \emph{NIPS}, 2012.

\bibitem[Ioffe and Szegedy(2015)]{ioffe2015batch}
Sergey Ioffe and Christian Szegedy.
\newblock Batch normalization: Accelerating deep network training by reducing
  internal covariate shift.
\newblock \emph{ICML}, 2015.

\bibitem[Sutskever et~al.(2014)Sutskever, Vinyals, and
  Le]{sutskever2014sequence}
Ilya Sutskever, Oriol Vinyals, and Quoc~V Le.
\newblock Sequence to sequence learning with neural networks.
\newblock In \emph{Advances in neural information processing systems}, pages
  3104--3112, 2014.

\bibitem[Laurent et~al.(2015)Laurent, Pereyra, Brakel, Zhang, and
  Bengio]{laurent2015batch}
C{\'e}sar Laurent, Gabriel Pereyra, Phil{\'e}mon Brakel, Ying Zhang, and Yoshua
  Bengio.
\newblock Batch normalized recurrent neural networks.
\newblock \emph{arXiv preprint arXiv:1510.01378}, 2015.

\bibitem[Amodei et~al.(2015)Amodei, Anubhai, Battenberg, Case, Casper,
  Catanzaro, Chen, Chrzanowski, Coates, Diamos, et~al.]{amodei2015deep}
Dario Amodei, Rishita Anubhai, Eric Battenberg, Carl Case, Jared Casper, Bryan
  Catanzaro, Jingdong Chen, Mike Chrzanowski, Adam Coates, Greg Diamos, et~al.
\newblock Deep speech 2: End-to-end speech recognition in english and mandarin.
\newblock \emph{arXiv preprint arXiv:1512.02595}, 2015.

\bibitem[Cooijmans et~al.(2016)Cooijmans, Ballas, Laurent, and
  Courville]{cooijmans2016recurrent}
Tim Cooijmans, Nicolas Ballas, C{\'e}sar Laurent, and Aaron Courville.
\newblock Recurrent batch normalization.
\newblock \emph{arXiv preprint arXiv:1603.09025}, 2016.

\bibitem[Salimans and Kingma(2016)]{salimans2016weight}
Tim Salimans and Diederik~P Kingma.
\newblock Weight normalization: A simple reparameterization to accelerate
  training of deep neural networks.
\newblock \emph{arXiv preprint arXiv:1602.07868}, 2016.

\bibitem[Neyshabur et~al.(2015)Neyshabur, Salakhutdinov, and
  Srebro]{neyshabur2015path}
Behnam Neyshabur, Ruslan~R Salakhutdinov, and Nati Srebro.
\newblock Path-sgd: Path-normalized optimization in deep neural networks.
\newblock In \emph{Advances in Neural Information Processing Systems}, pages
  2413--2421, 2015.

\bibitem[Amari(1998)]{amari1998natural}
Shun-Ichi Amari.
\newblock Natural gradient works efficiently in learning.
\newblock \emph{Neural computation}, 1998.

\bibitem[Vendrov et~al.(2016)Vendrov, Kiros, Fidler, and
  Urtasun]{vendrov2015order}
Ivan Vendrov, Ryan Kiros, Sanja Fidler, and Raquel Urtasun.
\newblock Order-embeddings of images and language.
\newblock \emph{ICLR}, 2016.

\bibitem[Team et~al.(2016)Team, Al-Rfou, Alain, Almahairi, Angermueller,
  Bahdanau, Ballas, Bastien, Bayer, Belikov, et~al.]{team2016theano}
The Theano~Development Team, Rami Al-Rfou, Guillaume Alain, Amjad Almahairi,
  Christof Angermueller, Dzmitry Bahdanau, Nicolas Ballas, Fr{\'e}d{\'e}ric
  Bastien, Justin Bayer, Anatoly Belikov, et~al.
\newblock Theano: A python framework for fast computation of mathematical
  expressions.
\newblock \emph{arXiv preprint arXiv:1605.02688}, 2016.

\bibitem[Lin et~al.(2014)Lin, Maire, Belongie, Hays, Perona, Ramanan,
  Doll{\'a}r, and Zitnick]{lin2014microsoft}
Tsung-Yi Lin, Michael Maire, Serge Belongie, James Hays, Pietro Perona, Deva
  Ramanan, Piotr Doll{\'a}r, and C~Lawrence Zitnick.
\newblock Microsoft coco: Common objects in context.
\newblock \emph{ECCV}, 2014.

\bibitem[Cho et~al.(2014)Cho, Van~Merri{\"e}nboer, Gulcehre, Bahdanau,
  Bougares, Schwenk, and Bengio]{cho2014learning}
Kyunghyun Cho, Bart Van~Merri{\"e}nboer, Caglar Gulcehre, Dzmitry Bahdanau,
  Fethi Bougares, Holger Schwenk, and Yoshua Bengio.
\newblock Learning phrase representations using rnn encoder-decoder for
  statistical machine translation.
\newblock \emph{EMNLP}, 2014.

\bibitem[Simonyan and Zisserman(2015)]{simonyan2014very}
Karen Simonyan and Andrew Zisserman.
\newblock Very deep convolutional networks for large-scale image recognition.
\newblock \emph{ICLR}, 2015.

\bibitem[Kiros et~al.(2014)Kiros, Salakhutdinov, and Zemel]{kiros2014unifying}
Ryan Kiros, Ruslan Salakhutdinov, and Richard~S Zemel.
\newblock Unifying visual-semantic embeddings with multimodal neural language
  models.
\newblock \emph{arXiv preprint arXiv:1411.2539}, 2014.

\bibitem[Kingma and Ba(2014)]{adam}
D.~Kingma and J.~L. Ba.
\newblock Adam: a method for stochastic optimization.
\newblock \emph{ICLR}, 2014.
\newblock arXiv:1412.6980.

\bibitem[Wang et~al.(2016)Wang, Li, and Lazebnik]{wang2015learning}
Liwei Wang, Yin Li, and Svetlana Lazebnik.
\newblock Learning deep structure-preserving image-text embeddings.
\newblock \emph{CVPR}, 2016.

\bibitem[Hermann et~al.(2015)Hermann, Kocisky, Grefenstette, Espeholt, Kay,
  Suleyman, and Blunsom]{hermann2015teaching}
Karl~Moritz Hermann, Tomas Kocisky, Edward Grefenstette, Lasse Espeholt, Will
  Kay, Mustafa Suleyman, and Phil Blunsom.
\newblock Teaching machines to read and comprehend.
\newblock In \emph{NIPS}, 2015.

\bibitem[Kiros et~al.(2015)Kiros, Zhu, Salakhutdinov, Zemel, Urtasun, Torralba,
  and Fidler]{kiros2015skip}
Ryan Kiros, Yukun Zhu, Ruslan~R Salakhutdinov, Richard Zemel, Raquel Urtasun,
  Antonio Torralba, and Sanja Fidler.
\newblock Skip-thought vectors.
\newblock In \emph{NIPS}, 2015.

\bibitem[Mikolov et~al.(2013)Mikolov, Chen, Corrado, and
  Dean]{mikolov2013efficient}
Tomas Mikolov, Kai Chen, Greg Corrado, and Jeffrey Dean.
\newblock Efficient estimation of word representations in vector space.
\newblock \emph{arXiv preprint arXiv:1301.3781}, 2013.

\bibitem[Zhu et~al.(2015)Zhu, Kiros, Zemel, Salakhutdinov, Urtasun, Torralba,
  and Fidler]{zhu2015aligning}
Yukun Zhu, Ryan Kiros, Rich Zemel, Ruslan Salakhutdinov, Raquel Urtasun,
  Antonio Torralba, and Sanja Fidler.
\newblock Aligning books and movies: Towards story-like visual explanations by
  watching movies and reading books.
\newblock In \emph{ICCV}, 2015.

\bibitem[Marelli et~al.(2014)Marelli, Bentivogli, Baroni, Bernardi, Menini, and
  Zamparelli]{marelli2014semeval}
Marco Marelli, Luisa Bentivogli, Marco Baroni, Raffaella Bernardi, Stefano
  Menini, and Roberto Zamparelli.
\newblock Semeval-2014 task 1: Evaluation of compositional distributional
  semantic models on full sentences through semantic relatedness and textual
  entailment.
\newblock \emph{SemEval-2014}, 2014.

\bibitem[Pang and Lee(2005)]{pang2005seeing}
Bo~Pang and Lillian Lee.
\newblock Seeing stars: Exploiting class relationships for sentiment
  categorization with respect to rating scales.
\newblock In \emph{ACL}, pages 115--124, 2005.

\bibitem[Hu and Liu(2004)]{hu2004mining}
Minqing Hu and Bing Liu.
\newblock Mining and summarizing customer reviews.
\newblock In \emph{Proceedings of the tenth ACM SIGKDD international conference
  on Knowledge discovery and data mining}, 2004.

\bibitem[Pang and Lee(2004)]{pang2004sentimental}
Bo~Pang and Lillian Lee.
\newblock A sentimental education: Sentiment analysis using subjectivity
  summarization based on minimum cuts.
\newblock In \emph{ACL}, 2004.

\bibitem[Wiebe et~al.(2005)Wiebe, Wilson, and Cardie]{wiebe2005annotating}
Janyce Wiebe, Theresa Wilson, and Claire Cardie.
\newblock Annotating expressions of opinions and emotions in language.
\newblock \emph{Language resources and evaluation}, 2005.

\bibitem[Gregor et~al.(2015)Gregor, Danihelka, Graves, and Wierstra]{draw}
K.~Gregor, I.~Danihelka, A.~Graves, and D.~Wierstra.
\newblock {DRAW}: a recurrent neural network for image generation.
\newblock arXiv:1502.04623, 2015.

\bibitem[Larochelle and Murray(2011)]{hugobmnist}
Hugo Larochelle and Iain Murray.
\newblock The neural autoregressive distribution estimator.
\newblock In \emph{AISTATS}, volume~6, page 622, 2011.

\bibitem[Liwicki and Bunke(2005)]{liwicki2005iam}
Marcus Liwicki and Horst Bunke.
\newblock Iam-ondb-an on-line english sentence database acquired from
  handwritten text on a whiteboard.
\newblock In \emph{ICDAR}, 2005.

\bibitem[Graves(2013)]{graves2013generating}
Alex Graves.
\newblock Generating sequences with recurrent neural networks.
\newblock \emph{arXiv preprint arXiv:1308.0850}, 2013.

\end{thebibliography}
